# Cross-Domain Multitask Learning with Latent Probit Models

**Shaobo Han, Xuejun Liao, Lawrence Carin**  SHAOBO.HAN,XJLIAO,LCARIN@DUKE.EDU
Duke University, Durham, NC 27519, USA

## Abstract

Learning multiple tasks across heterogeneous domains is a challenging problem since the feature space may not be the same for different tasks. We assume the data in multiple tasks are generated from a latent common domain via sparse domain transforms and propose a latent probit model (LPM) to jointly learn the domain transforms, and a probit classifier shared in the common domain. To learn meaningful task relatedness and avoid over-fitting in classification, we introduce sparsity in the domain transforms matrices, as well as in the common classifier parameters. We derive theoretical bounds for the estimation error of the classifier parameters in terms of the sparsity of domain transform matrices. An expectation-maximization algorithm is derived for learning the LPM. The effectiveness of the approach is demonstrated on several real datasets.

## 1. Introduction

There are two basic approaches for analysis of data from two or more tasks, single-task learning (STL) and multi-task learning (MTL). Whereas STL solves each task in isolation, with possible relations between the tasks ignored, MTL solves the tasks jointly, exploiting between-task relations to reduce the hypothesis space and improve generalization (Baxter, 2000). The advantage of MTL is known to be manifested when the tasks are truly related and the task relations are appropriately employed. For supervised learning, in particular, MTL can achieve the same level of generalization performance as STL, and yet uses significantly fewer labeled examples per task (Baxter, 2000). The reduced sample complexity in each task is achieved by transferring labeling information from related tasks.

While the MTL literature has primarily assumed that the tasks have the same input and output domains and differ only in data distributions (Baxter, 2000; Bakker & Heskes, 2003; Argyriou et al., 2007; Ben-David & Borbely, 2008), a number of recent publications are beginning to break the limit of this assumption, in an attempt of extending MTL to a wider range of applications (He & Rick, 2011; Maayan & Mannor, 2011; Kulis et al., 2011; Wang & Mahadevan, 2011).

In these recent publications, different tasks are permitted to have different feature spaces. In particular, (He & Rick, 2011) simultaneously performs multi-view learning in each task and multi-task learning in shared views, assuming each task has its own features but may also share features with other tasks. The method in (Maayan & Mannor, 2011) allows tasks to have different feature representations, learning rotations between the feature representations by matching the tasks' empirical means and covariance matrices. The work in (Kulis et al., 2011) considers a source task and a target task, assumed to have different feature dimensions, and learns a nonlinear transformation between the source feature domain and the target feature domain using kernel techniques. Finally, (Wang & Mahadevan, 2011) employs a manifold alignment technique to map each task's input domain to a common latent space, with the task-specific maps achieving the goal of simultaneously clustering examples with the same label, separating examples with different labels, and preserving the topology of each task's manifold.

In this paper, we address the problem of multi-task learning across heterogenous domains, assuming that each task is a binary classification with a task-specific feature representation. The approach we take differs from (Maayan & Mannor, 2011; Kulis et al., 2011; Wang & Mahadevan, 2011) in several important aspects. First, while these previous methods all learn domain transforms and classification in two separate steps, we integrate the two steps by learning domain transforms and classification jointly. Secondly, the domain transforms in our approach are represented by sparse matrices, with the sparsity enforced by a Lapla-cian prior on the transform matrices (corresponding to an $\ell_1$ penalty to the log-likelihood). By contrast, all previous methods do not impose sparsity on domain transforms. The third difference is that the overall model in our approach consists of a factor model for the observed features, which can be used to synthesize new data unseen during training. Finally, our approach is semi-supervised, using labeled as well as unlabeled examples to jointly find the domain transforms and the classification. By contrast, the methods in (Maayan & Mannor, 2011; Kulis et al., 2011) are supervised, and the method in (Wang & Mahadevan, 2011) is semi-supervised in learning domain transforms, but supervised in learning classification. While full supervision can be challenged by the scarcity of labeled examples (typically assumed in MTL), semi-supervision is doubly beneficial to a joint learning approach, in which unlabeled examples help to perform classification, while labeled examples help to find the domain transforms.

The proposed approach is based on a sparse hierarchical Bayesian model, referred to as the *latent probit model* (LPM), which jointly represents the sparse domain transforms and a common sparse probit classifier (Albert & Chib, 1993) in the latent feature space, with the sparsity imposed by a hierarchical Laplacian prior (Figueiredo, 2003). We employ expectation-maximization (EM) to find the maximum *a posterior* (MAP) solution to the domain transforms and probit parameters.

The sparsity of domain transforms in LPM plays a pivotal role in defining the between-task relations. Roughly speaking, a greater sparsity in domain transforms indicates closer relations between the tasks. In other words, sparser domain transforms imply that different tasks look more similar to each other in the latent feature space, and thus greater performance gain may be achieved by sharing information among the tasks. We give a quantitative analysis of the performance gain by providing an upper bound to the estimation error of the probit classifier, which is shared among the tasks in the latent space. The bound has an analytic functional dependence on the sparsity level of domain transforms, showing that sparsity contributes directly to the error reduction. In addition, the bound also reveals the error's dependency on the number of tasks, the number of labeled examples in each task, and the latent dimensionality.

## 2. The Latent probit Model

The latent probit model (LPM) is a generative probabilistic model for $M \geq 2$ partially labeled sets of feature vectors (data points), assuming each dataset has a different feature representation. The LPM has a hierarchical Bayesian structure, as graphically shown in Figure 1, and is parameterized by $\{\eta, \boldsymbol{\mu}, \boldsymbol{\Sigma}, b, \mathbf{w}\}$ and $\{\mathbf{F}_m, \mathbf{d}_m\}_{m=1}^M$. The parameters $\mathbf{w}$ specify the probit classifier shared by the tasks in the latent feature space, and $\mathbf{F}_m$ specifies the domain transform for the $m$-th dataset up to a translation (which is specified by $\mathbf{d}_m$). The parameters $\mathbf{w}$ and $\{\mathbf{F}_m\}_{m=1}^M$ are given hierarchical Laplacian priors (Figueiredo, 2003) to encourage sparsity, with the priors specified by hyperparameters $\{\gamma, \lambda\}$. The other hollow circles in Figure 1 denote latent variables, which include $\{\boldsymbol{\tau}, \mathbf{u}, \mathbf{s}, z\}$. The generative process in the LPM is described below, with $\mathcal{N}(\boldsymbol{\mu}, \boldsymbol{\Sigma})$ denoting a normal distribution with mean $\boldsymbol{\mu}$ and covariance matrix $\boldsymbol{\Sigma}$.

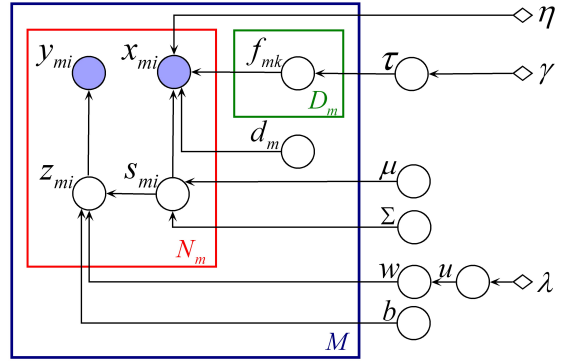

*Figure 1.* A graphic representation of the proposed latent probit model, where solid circles denote data, hollow circles denote unknown parameters and latent variables, and diamonds denote input parameters (including hyperparameters and fixed model parameters).

Given hyper-parameters $\{\gamma, \lambda\}$, the sparse parameters $\mathbf{w}$ and $\{\mathbf{F}_m\}_{m=1}^M$ are generated as follows.

1. Draw $\mathbf{w} = [w_j]_{F_0 \times 1}$, the sparse parameters of the probit model shared by the tasks in the latent feature space,

$$\begin{aligned} w_j &\sim \mathcal{N}(0, u_j), \\ u_j &\sim \frac{\lambda}{2} \exp\{-\frac{\lambda}{2} u_j\}, \ u_j \geq 0, \ j = 1, 2, ..., F_0, \end{aligned}$$

where $F_0$ is the latent feature dimensionality.

2. For $m = 1, 2, \cdots, M$, draw the sparse domain transform matrix $\mathbf{F}_m = [f_{mkj}]_{D_m \times F_0}$ by

$$\begin{aligned} f_{mkj} &\sim \mathcal{N}(0, \tau_{mkj}), \\ \tau_{mkj} &\sim \frac{\gamma}{2} \exp\{-\frac{\gamma}{2} \tau_{mkj}\}, \ \tau_{mkj} \geq 0, \end{aligned}$$

$k = 1, \cdots, D_m$ and $j = 1, ..., F_0$, with $D_m$ the observed feature dimensionality of the $m$-th dataset.

Given parameters $\{\eta, \boldsymbol{\mu}, \boldsymbol{\Sigma}, b, \mathbf{w}\} \cup \{\mathbf{F}_m, \mathbf{d}_m\}_{m=1}^M$, the data sets are generated as follows.

For $i = 1, 2, \cdots, N_m$ and $m = 1, 2, \cdots, M$,

1. Draw a latent feature vector

$$\mathbf{s}_{mi} \sim \mathcal{N}(\boldsymbol{\mu}, \boldsymbol{\Sigma}), \qquad (1)$$

where $\boldsymbol{\mu} \in \mathbb{R}^{F_0 \times 1}$ and $\boldsymbol{\Sigma} \in \mathbb{R}^{F_0 \times F_0}$ are the mean and covariance matrix, respectively.

2. Draw an observed feature vector

$$\mathbf{x}_{mi} \sim \mathcal{N}(\mathbf{F}_m \mathbf{s}_{mi} + \mathbf{d}_m, \eta \mathbf{I}), \qquad (2)$$

where $\mathbf{d}_m \in \mathbb{R}^{D_m}$, $\eta > 0$ and $\mathbf{I}$ denotes an identity matrix of appropriate dimensions.

3. If the feature vector $\mathbf{x}_{mi}$ requires a label, draw the label by

$$\begin{aligned} y_{mi} &= \begin{cases} +1, & \text{if } z_{mi} \geq 0, \\ -1, & \text{otherwise}, \end{cases} \\ z_{mi} &\sim \mathcal{N}(\mathbf{w}^T \mathbf{s}_{mi} + b, 1), \quad b \in \mathbb{R}. \end{aligned} \qquad (3)$$

Note that the latent normal distribution in (1) can be extended to a mixtures of normal distributions to account for more complicated data manifolds.

With $\{\mathbf{F}_m\}_{m=1}^M$ drawn from sparse prior distributions, most entries of these matrices will be zero; by (2) this implies that only a few latent features are responsible for generating the observed features. Since this is true for any $m$, the chance for different datasets to use the same features to generate their observed features is large. However, latent features are identically distributed; thus the shared latent features must have the same statistics across the tasks. Therefore, the datasets (sets of features vectors) generated by the LPM model are encouraged to be closely related.

While the sparsity of $\{\mathbf{F}_m\}_{m=1}^M$ reflects the relatedness between the sets of features vectors, the sparsity of $\mathbf{w}$ encourages the classification to be dependent on a few latent features. This is important, because even when the observed features differ among tasks to entail less sparse $\{\mathbf{F}_m\}_{m=1}^M$, the tasks may still be able to share information for classification through appropriately selected latent features.

## 3. Theoretical Analysis of the LPM

The goal of our analysis is to quantify the notion that sparse domain transforms encourage the tasks to be related, and that better generalization can be achieved by sharing information among related tasks to learn the common classifier. The analysis is based on an upper bound for the estimation error of $\mathbf{w}$, with the bound represented in terms of the number of nonzero elements of the true $\{\mathbf{F}_m\}_{m=1}^M$.

Since we are analyzing the general information-sharing mechanism in the LPM, we expect the results to be insensitive to the choice of estimation method. We therefore employ a simple two-step approach to estimate $\mathbf{w}$. The estimation is based on training data generated by the true LPM parameterized by $\{\eta, \boldsymbol{\mu}, \boldsymbol{\Sigma}, b, \mathbf{w}^*\} \cup \{\mathbf{F}_m, \mathbf{d}_m\}_{m=1}^M$, with the simplifications $b = 0$, $\boldsymbol{\Sigma} = \mathbf{I}$, $\boldsymbol{\mu} = \mathbf{0}$, and $\mathbf{d}_m = \mathbf{0}\ \forall m$, where $\mathbf{0}$ is a vector of zeros of appropriate dimensions. Note we have used a superscript $*$ to emphasize $\mathbf{w}^*$ is the vector of unknown parameters to be estimated.

Let $\{\mathbf{X}_m\}_{m=1}^M$, with $\mathbf{X}_m = [\mathbf{x}_{m1}, \mathbf{x}_{m2}, \cdots, \mathbf{x}_{mL_m}]$, be $M$ sets of feature vectors, each corresponding to a task. By the generative process of the LPM,

$$\mathbf{X}_m = \mathbf{F}_m \mathbf{S}_m + [\epsilon_{mij}]_{D_m \times N_m},$$

where $\{\epsilon_{mij}\}$ are i.i.d. drawn from a zero-mean normal distribution with variance $\eta$, and the entries of $\mathbf{S}_m$ are i.i.d. from the standard normal distribution. Given $\mathbf{X}_m$, the maximum-likelihood solutions to $\{\mathbf{S}_m\}$ are given by

$$\widehat{\mathbf{S}}_m = (\mathbf{F}_m^T \mathbf{F}_m)^{-1} \mathbf{F}_m^T \mathbf{X}_m, \ \forall m, \qquad (4)$$

which form a global data matrix by pooling data across the tasks,

$$\boldsymbol{\Psi} = [\widehat{\mathbf{S}}_1, \widehat{\mathbf{S}}_2, ..., \widehat{\mathbf{S}}_M] \in \mathbb{R}^{F_0 \times n_t}, \qquad (5)$$

where $n_t = \sum_{m=1}^M L_m$ is the total number of training examples across all $M$ tasks.

To simplify the analysis, we assume access to the latent responses of $\mathbf{w}^*$ to $\boldsymbol{\Psi}$, i.e,

$$\mathbf{z} = \boldsymbol{\Psi}^T \mathbf{w}^* + \mathbf{e} \qquad (6)$$

where $\mathbf{z} = [\mathbf{z}_1, \cdots, \mathbf{z}_M]^T$ with $\mathbf{z}_m = [z_{m1}, \cdots, z_{mL_m}]$, and the entries in $\mathbf{e}$ are assumed i.i.d. from the standard normal distribution. These assumptions may be avoided at the price of complicating the bound, which is not pursued here. The estimate of $\mathbf{w}^*$ is given by

$$\widehat{\mathbf{w}} = \arg\min_{\mathbf{w}} \left( \|\mathbf{z} - \boldsymbol{\Psi}^T \mathbf{w}\|_2^2 + r\|\mathbf{w}\|_1 \right). \qquad (7)$$

We derive an upper bound to $\|\widehat{\mathbf{w}} - \mathbf{w}^*\|_2$, following similar arguments as in (Bickel et al., 2009; Lounici et al., 2009) and making use of a key result in (Byrne, 2009) on extreme singular values of Hermitian matrices. Our main results are stated in Theorem 1, the proof of which is in the Appendix.

**Theorem 1.** *Let $\mathbf{w}^*$ have nonzero and zero elements indexed respectively by $J$ and $J^c$. Denote $s = |J|$ as the cardinality of $J$. Let $\boldsymbol{\delta} = \widehat{\mathbf{w}} - \mathbf{w}^*$ with $\widehat{\mathbf{w}}$ given in (7), and $c_0$ be the minimum nonnegative number such that $\|\boldsymbol{\delta}_{J^c}\|_1 \leq c_0 \|\boldsymbol{\delta}_J\|_1$. Let $\boldsymbol{\psi}_j$ be the transpose of the $j$-th row of $\boldsymbol{\Psi}$ and $\varepsilon_\psi = \max_j \|\boldsymbol{\psi}_j\|_2$. For any $F_0 \geq 2$ and $a \geq \sqrt{8}$, it holds with probability of at least $P_e = 1 - F_0^{1-a^2/8}$ that*

$$\|\boldsymbol{\delta}\|_2 \leq \frac{2a\varepsilon_\psi n_t^{-1}\sqrt{s(1+c_0^2 s)\ln F_0}}{\sum_{m=1}^M \frac{\omega_{\min}(\mathbf{X}_m^T \mathbf{X}_m/n_t)}{\max_i \left(\sum_{j=1}^{F_0} \|\mathbf{f}_{m,:,j}\|_0 |f_{ij}|^2\right)}}, \quad (8)$$

*where $\mathbf{f}_{m,:,j}$ denotes the $j$-th column of $\mathbf{F}_m$ and $\|\mathbf{f}\|_0$ denotes the number of nonzero elements in vector $\mathbf{f}$.*

The bound in (8) establishes the functional dependency of $\|\widehat{\mathbf{w}} - \mathbf{w}^*\|_2$ on a number of characteristic parameters of the LPM. Foremost, the term $\|\mathbf{f}_{m,:,j}\|_0$ measures the number of nonzero elements in the $j$-th column of $\mathbf{F}_m$. A sparse $\mathbf{F}_m$ has small $\|\mathbf{f}_{m,:,j}\|_0$ for its columns, which decreases the term $\max_i \left(\|\mathbf{f}_{m,:,j}\|_0 \sum_{j=1}^{F_0} |f_{ij}|^2\right)$ and contributes to the error reduction. Second, $s$ is the number of nonzero elements in $\mathbf{w}$; a sparse $\mathbf{w}$ has a small $s$, which makes the error small.

Third, recall that $n_t = \sum_{m=1}^M L_m$, where $M$ is the number of tasks, and $L_m$ is the number of training samples in the $m$-th task. The $n_t$ in the denominator of (8) plays the role of normalization with respect to the training examples across all tasks, leaving the $n_t$ in the numerator to influence the error: large $n_t$ indicates small error. Note that some tasks may have few examples while other have abundant ones; as long as they add up to a large $n_t$, similar error reduction will be achieved. Lastly, $F_0$ is the dimensionality of latent features shared across the tasks. The error bound decreases as $F_0$ becomes smaller.

## 4. Parameter Estimation

We seek a MAP estimate of the parameters $\Theta = \{\boldsymbol{\mu}, \boldsymbol{\Sigma}, b, \mathbf{w}\} \cup \{\mathbf{F}_m, \mathbf{d}_m\}_{m=1}^M$. Taking into account all data (labeled and unlabeled) and the sparse priors, and integrating out the latent variables $\{\boldsymbol{\tau}, \mathbf{u}, \mathbf{s}, z\}$, one obtains the logarithmic posterior probability of $\Theta$,

$$\ell(\Theta) = \sum_{m=1}^M \sum_{i \in \mathcal{U}_m} \ln \int p(\mathbf{x}_{mi}, \mathbf{s}_{mi}|\Theta) d\mathbf{s}_{mi}$$
$$+ \sum_{m=1}^M \sum_{i \in \mathcal{L}_m} \ln \int p(\mathbf{x}_{mi}, y_{mi}, z_{mi}, \mathbf{s}_{mi}|\Theta) dz_{mi} d\mathbf{s}_{mi}$$
$$+ \sum_{m=1}^M \sum_k \sum_j \ln \int p(f_{mkj}|\tau_{mkj}) p(\tau_{mkj}|\gamma) d\tau_{mkj},$$
$$+ \sum_j \ln \int p(w_j|u_j) p(u_j|\lambda) du_j$$

labeled and unlabeled feature vectors in the $m$-th data set, i.e., $\mathcal{L}_m \cup \mathcal{U}_m = \{1, 2, \cdots, N_m\}$.

We employ an expectation-maximization (EM) algorithm to maximize $\ell(\Theta)$, with $\{\eta, F_0\}$ and hyperparameters $\{\gamma, \lambda\}$ treated as input parameters to the algorithm, determined separately by cross-validation when necessary. The EM algorithm consists of an iteration of E-step and M-step. In the E-step, one computes the conditional moments of latent variables $\{z_{mi}, \mathbf{s}_{mi}, \boldsymbol{\tau}, \mathbf{u}\}$ given the data and the most recent parameters $\Theta$. In the M-step, one calculates the updated model parameters $\widehat{\Theta}$ using the latent variables' moments obtained in E-step. The complete EM algorithm is given in Algorithm 1, with major update equations summarized below. The algorithm requires $O(F_0 \sum_{m=1}^M D_m(F_m + F_0^2))$ scalar products per iteration.

**Update of Latent Features' Distribution**[1]

$$\widehat{\boldsymbol{\mu}} = \frac{1}{n_a} \sum_{m=1}^M \sum_{i=1}^{N_m} \boldsymbol{\phi}_{mi} \quad (9a)$$

$$\widehat{\boldsymbol{\Sigma}} = \frac{1}{n_a} \sum_{m=1}^M \sum_{i=1}^{N_m} \left((\boldsymbol{\phi}_{mi} - \boldsymbol{\mu})(\boldsymbol{\phi}_{mi} - \boldsymbol{\mu})^T + \mathbf{R}_m \mathbf{w} \beta_{mi} \mathbf{w}^T \mathbf{R}_m + \mathbf{R}_m\right) \quad (9b)$$

where $n_a = \sum_{m=1}^M N_m$.

$\boldsymbol{\phi}_{mi} = \mathbf{R}_m(\boldsymbol{\Sigma}^{-1}\boldsymbol{\mu} + \mathbf{w}(\xi_{mi} - b) + \eta^{-1}\mathbf{F}_m^T(\mathbf{x}_{mi} - \mathbf{d}_m))$,
$\mathbf{R}_m = (\boldsymbol{\Sigma}^{-1} + \mathbf{w}\mathbf{w}^T + \eta^{-1}\mathbf{F}_m^T\mathbf{F}_m)^{-1}$,
$\beta_{mi} = \begin{cases} \rho_{mi}, & \text{if } i \in \mathcal{U}_m, \\ (\zeta_{mi}^2 + \rho_{mi})g_{cdf}(\frac{\zeta_{mi}}{\sqrt{\rho_{mi}}}) \\ \quad + \zeta_{mi}\sqrt{\rho_{mi}}g_{pdf}(\frac{\zeta_{mi}}{\sqrt{\rho_{mi}}}), & \text{if } i \in \mathcal{L}_m, \end{cases}$
$\xi_{mi} = \begin{cases} \zeta_{mi}, & \text{if } i \in \mathcal{U}_m, \\ \zeta_{mi}g_{cdf}(\frac{\zeta_{mi}}{\sqrt{\rho_{mi}}}) \\ \quad + \sqrt{\rho_{mi}}g_{pdf}(\frac{\zeta_{mi}}{\sqrt{\rho_{mi}}}), & \text{if } i \in \mathcal{L}_m, \end{cases}$
$\rho_{mi} = 1 + \mathbf{w}^T\mathbf{Q}_m\mathbf{w}$,
$\zeta_{mi} = \mathbf{w}^T\boldsymbol{\mu} + b + \eta^{-1}\mathbf{w}^T\mathbf{Q}_m\mathbf{F}_m^T(\mathbf{x}_{mi} - \mathbf{F}_m\boldsymbol{\mu} - \mathbf{d}_m)$,
$\mathbf{Q}_m = (\boldsymbol{\Sigma}^{-1} + \eta^{-1}\mathbf{F}_m^T\mathbf{F}_m)^{-1}$.

**Update of Domain Transforms**

$$\widehat{\mathbf{d}}_m = \frac{1}{N_m}\sum_{i=1}^{N_m}(\mathbf{x}_{mi} - \mathbf{F}_m\boldsymbol{\phi}_{mi}), \quad (10a)$$

$$\widehat{\mathbf{f}}_{mk} = \mathbf{V}_{mk}(\alpha\mathbf{I}_{F_0} + \mathbf{V}_{mk}\boldsymbol{\Gamma}_{m1}\mathbf{V}_{mk})^{-1}\mathbf{V}_{mk}$$
$$\times \sum_{i=1}^{N_m}\boldsymbol{\phi}_{mi}^T(x_{mik} - \widehat{d}_{mk}), \quad (10b)$$

---

[1] It can be shown that, under the LPM, the marginal distribution of $\mathbf{x}_{mi}$ is $\mathcal{N}(\mathbf{F}_m\boldsymbol{\mu} + \mathbf{d}_m, \eta\mathbf{I} + \mathbf{F}_m\boldsymbol{\Sigma}\mathbf{F}_m^T)$, with the mean and covariance matrix defined duplicately by $(\boldsymbol{\mu}, \mathbf{d}_m)$ and $(\mathbf{F}_m, \boldsymbol{\Sigma})$, respectively. Similar situations exist for $z_{mi}$. To void duplicatedness, one may wish to set $\boldsymbol{\mu} = \mathbf{0}$, $\boldsymbol{\Sigma} = \mathbf{I}$, and do not update them during learning.

**Algorithm 1** The EM algorithm for learning the LPM
- **Input:** $\{\mathbf{x}_{mi}\}_{i=1}^{N_m} \cup \{y_{mi}\}_{i\in\mathcal{L}_m}$, $m=1,2,\cdots,M$; $\{\gamma,\lambda\}$ and $\{\eta,F_0\}$.
- **Initialize** $\Theta$.
- **repeat**
  - Update $\boldsymbol{\Sigma}, \boldsymbol{\mu}$ using $\{\mathbf{x}_{mi}\}_{i=1}^{N_m} \cup \{y_{mi}\}_{i\in\mathcal{L}_m}$, $m=1,2,\cdots,M$, according to (9)
  - **for** $m=1$ **to** $M$ **do**
    - Update $\mathbf{F}_m, \mathbf{d}_m$ using $\{\mathbf{x}_{mi}\}_{i=1}^{N_m} \cup \{y_{mi}\}_{i\in\mathcal{L}_m}$ according to (10)
  - **end for**
  - Estimate $\mathbf{w}, b$ according to (11) using $\{\mathbf{x}_{mi}\}_{i\in\mathcal{L}_m} \cup \{y_{mi}\}_{i\in\mathcal{L}_m}$, $m=1,2,\cdots,M$,
- **until** $\ell(\Theta)$ Converges

for $k=1,2,\cdots,F_0$ and $m=1,2,\cdots,M$, where $\alpha = \eta\sqrt{\gamma}$ is a regularization parameter and

$$\boldsymbol{\Gamma}_{m1} = \sum_{i=1}^{N_m}(\boldsymbol{\phi}_{mi}\boldsymbol{\phi}_{mi}^T + \mathbf{R}_m + \beta_{mi}\mathbf{R}_m\mathbf{w}\mathbf{w}^T\mathbf{R}_m),$$
$$\mathbf{V}_{mk} = \mathrm{diag}(\sqrt{|f_{mk1}|}, \sqrt{|f_{mk2}|}, \cdots, \sqrt{|f_{mkF_0}|}).$$

**Update of Probit Classifier**

$$\widehat{\mathbf{w}} = \mathbf{G}(\vartheta\mathbf{I}+\mathbf{G}\boldsymbol{\Gamma}_1\mathbf{G})^{-1}\mathbf{G}\sum_{m=1}^{M}\sum_{i\in\mathcal{L}_m}\boldsymbol{\phi}_{mi}(\xi_{mi}-b), \quad (11a)$$

$$\widehat{b} = \frac{1}{\sum_{m=1}^{M}N_m}\sum_{m=1}^{M}\sum_{i\in\mathcal{L}_m}(\xi_{mi} - \boldsymbol{\phi}_{mi}^T\widehat{\mathbf{w}}), \quad (11b)$$

where $\vartheta = \sqrt{\lambda}$ is another regularization parameter and

$$\boldsymbol{\Gamma}_2 = \sum_{m=1}^{M}\sum_{i\in\mathcal{L}_m}(\boldsymbol{\phi}_{mi}\boldsymbol{\phi}_{mi}^T + \mathbf{R}_m + \mathbf{R}_m\mathbf{w}\beta_{mi}\mathbf{w}^T\mathbf{R}_m),$$
$$\mathbf{G} = \mathrm{diag}(\sqrt{|w_1|}, \sqrt{|w_2|}, \cdots, \sqrt{|w_{F_0}|}).$$

## 5. Experimental Results

### 5.1. Cancer Diagnosis

We first consider the two Wisconsin breast cancer datasets (original and diagnostic) from the UCI machine learning repository[2]. The objective of both tasks is to identify benign or malignant cells. The feature dimensionality is 9 for the original data and 30 for the diagnostic data. We set $F_0$ to the smallest dimensionality among the tasks to favor error reduction (as suggested by (8)), and $\eta = 10^{-3}$ to enlarge the role of domain transforms in connecting the tasks, with the regularization parameters $(\alpha, \vartheta)$ determined via cross-validation (the robustness to these parameters is shown below). We perform both multitask learning and transfer learning experiments, and compare the LPM to STL and the methods in (Wang & Mahadevan, 2011) (abbreviated as HDAMA), (Maayan & Mannor, 2011), and (Kulis et al., 2011), with all competing methods using standard probit classifiers. The method in (Maayan & Mannor, 2011) cannot perform MTL and is excluded in the comparisons on MTL. The performance is measured in terms of the area under ROC curve (AUC), as a function of the number of labeled examples per task in the MTL case, or the number of labeled examples in the target task in the transfer learning case. The results are averaged over 50 independent runs, each constituting an independent split of the data into training sets and test sets.

Figure 2(a) shows that, for MTL, the LPM performs comparably as or slightly better than HDAMA and both outperform the other methods, especially when labeled data are scarce. In transfer learning, all data in the source domain are labeled, and we have only a few labeled data in the target domain. We transfer all the labeled data from the source domain to the target domain. Figure 2(b-c) show that the performance of the LPM is slightly better than HDAMA, probably due to the fact that the amount of data (labeled and unlabeled) is balanced between the two tasks.

The regularization parameters $\alpha$ and $\vartheta$ control the sparsity of domain transforms and the classifier, respectively. Table 1 summarizes the performance of the LPM relative to STL, under a wide range of settings for these parameters. The importance of sparsity is indicated by the diminishing performance improvements as the regularization parameters approach zero. Over a wide range in the middle, the LPM maintains stable performance improvements over STL, indicating the learning is robust to the settings of regularization parameters. The table also shows that the sparsity of domain transforms plays a more prominent role in influencing the performance than the classifier itself, signaling that the benefit of sharing information among the tasks can outweigh the benefit of feature selection.

---
[2]UC Irvine Machine Learning Repository: http://archive.ics.uci.edu/ml/.

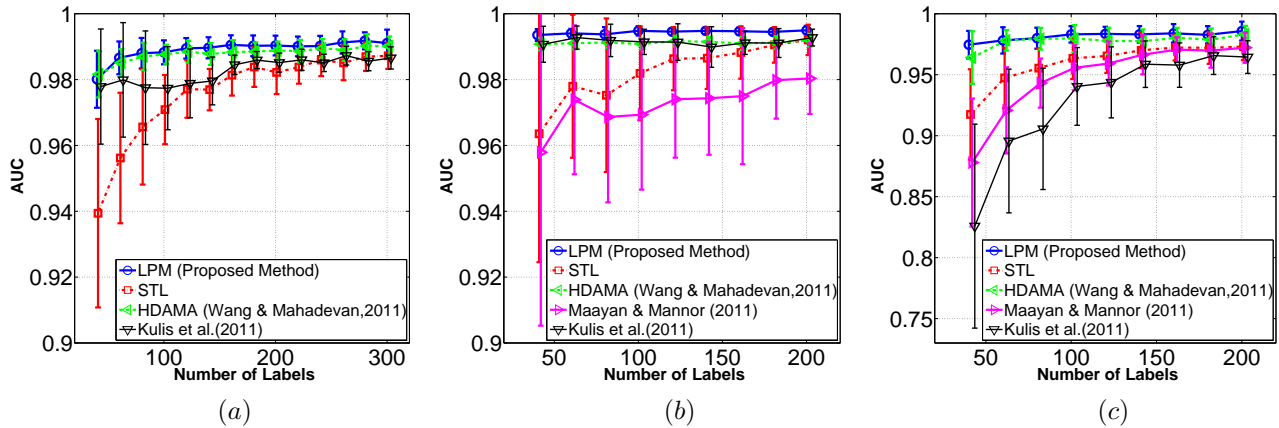

*Figure 2.* A comparison of performance on the Wisconsin breast cancer data; (a) multitask learning; (b) transfer learning with the original data as the source domain and the diagnostic data as the target domain; (c) transfer learning with the diagnostic data as the source domain and the original data as the target domain.

*Table 1.* The performance of the LPM on Wisconsin breast cancer data with $\alpha = \eta\sqrt{\gamma}$ and $\vartheta = \sqrt{\lambda}$ taking various values. The numbers shown are the improvements in AUC (%) relative to STL, averaged over 50 independent runs.

|  | LABEL=50 | | | | LABEL=100 | | | | LABEL=150 | | | |
| --- | --- | --- | --- | --- | --- | --- | --- | --- | --- | --- | --- | --- |
|  | $\vartheta=0$ | $\vartheta=0.1$ | $\vartheta=1$ | $\vartheta=10$ | $\vartheta=0$ | $\vartheta=0.1$ | $\vartheta=1$ | $\vartheta=10$ | $\vartheta=0$ | $\vartheta=0.1$ | $\vartheta=1$ | $\vartheta=10$ |
| $\alpha=0$ | 1.69 | 1.69 | 1.74 | 1.81 | -0.28 | -0.27 | -0.24 | -0.17 | -0.68 | -0.67 | -0.64 | -0.58 |
| $\alpha=0.01$ | 2.40 | 2.37 | 2.40 | 2.12 | 0.51 | 0.52 | 0.57 | 0.49 | 0.20 | 0.20 | 0.23 | 0.19 |
| $\alpha=0.05$ | 2.42 | 2.43 | 2.51 | 1.95 | 0.67 | 0.68 | 0.73 | 0.66 | 0.36 | 0.36 | 0.39 | 0.38 |
| $\alpha=0.1$ | 2.42 | 2.44 | 2.55 | 1.97 | 0.74 | 0.75 | 0.79 | 0.74 | 0.40 | 0.40 | 0.43 | 0.44 |
| $\alpha=0.5$ | 2.35 | 2.37 | 2.49 | 2.09 | 0.70 | 0.71 | 0.74 | 0.66 | 0.42 | 0.43 | 0.46 | 0.48 |
| $\alpha=1$ | 1.73 | 1.78 | 1.95 | 1.34 | 0.40 | 0.42 | 0.49 | 0.55 | 0.26 | 0.27 | 0.29 | 0.27 |
| $\alpha=5$ | 2.51 | 2.51 | 2.57 | 2.00 | 0.74 | 0.74 | 0.79 | 0.72 | 0.41 | 0.41 | 0.43 | 0.42 |
| $\alpha=10$ | -1.36 | -1.17 | -0.90 | -0.46 | -2.23 | -2.13 | -2.03 | -1.68 | -1.71 | -1.70 | -1.66 | -1.56 |

### 5.2. Mine Detection

The land-mine detection problem (Xue et al., 2007) is based on airborne synthetic-aperture radar (SAR) data and the underwater mine detection problem (Liu et al., 2009) is based on synthetic-aperture sonar (SAS) data[3]. Here we solve these two problems together, using the proposed cross-domain multitask learning approach. The feature dimensionality of land-mine data is 9 and that of underwater mine data is 13, and the labels do not have the same exact meaning for the two problem domains. There are a total of 19 land-mine tasks and 8 underwater mine tasks. The number of data points in the underwater mine tasks ranges from 756 to 3562, which is much larger than that for the land-mine tasks (ranging from 445 to 454). This problem can be viewed as a multitask learning across heterogeneous input and output domains (although the labels have known correspondence). We consider 9 land-mine tasks and all 8 underwater tasks, pairing them up to form $9 \times 8 = 72$ MTL problems. The results are reported as an average over the 72 problems, with the setting of $F_0$ and regularization parameters based on the same rule as in Section 5.1.

The performance comparisons for multi-task learning are shown in Figure 3(a) in terms of average AUC. Each curve results from an average of 100 independent runs of independently splitting the data into training and test sets and $9 \times 8$ combinations of underwater tasks versus land-mine tasks. In the transfer learning case, 50 labeled samples together with all other unlabeled samples are transferred to the target domain. The performance on the target task is shown in Figure 3(b-c). It is seen that the LPM outperforms all other methods by significantly large margins, in both multi-task learning and transfer learning from landmine data to underwater mine data. The competition on transfer learning from underwater mine data to land-mine data is more intense, but the LPM still gives the best overall outperformance.

While the amount of examples is balanced between the two Wisconsin tasks, it is highly unbalanced between the land-mine tasks and the underwater mine tasks (as detailed above). The results indicate that the LPM is more robust to this unbalance than the other methods.

---
[3]The land-mine data are available at http://www.ee.duke.edu/~lcarin/LandmineData.zip and the underwater mine data are available at http://www.ece.duke.edu/~lcarin/UnderwaterMines.zip

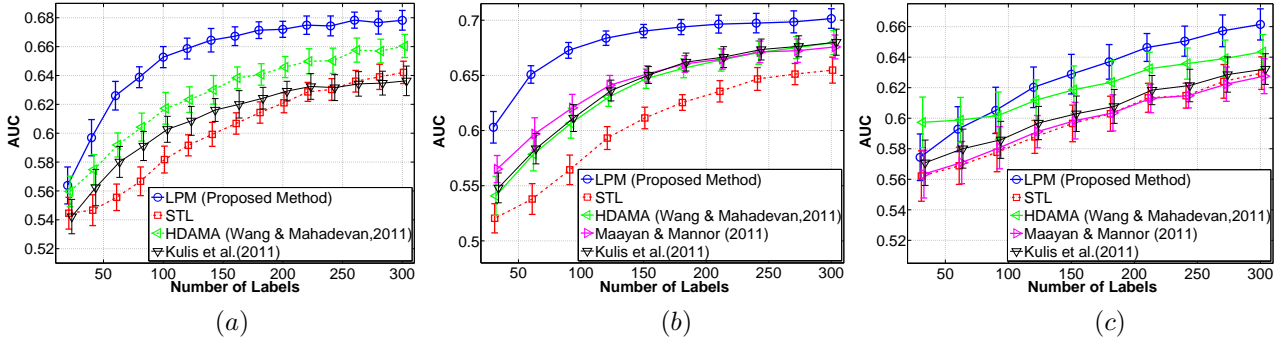

Figure 3. A comparison of performance on the land-mine/underwater mine detection problem; (a) multitask learning; (b) transfer learning with land-mine data as the source domain and underwater mine data as the target domain; (c) transfer learning with underwater mine data as the source domain and land-mine data as the target domain.

## 6. Conclusions

We have proposed the LPM model for cross-domain multi-task learning, assuming heterogenous feature representations across the tasks. The benefit of MTL in the LPM is based on the tasks' relatedness in the latent feature space, which is characterized by the sparse domain transforms. By promoting sparseness of domain transforms and the common classifiers, information sharing is encouraged to the advantage of improving performance in each individual task. The importance of sparsity is demonstrated by both theoretical analysis and experimental results.

## Acknowledgement

The research reported here has been supported by the ONR ATL program.

## Appendix

**Proof of Theorem 1**

By (7), one has

$$\frac{1}{n_t}\|\boldsymbol{\Psi}^T\widehat{\mathbf{w}}-\mathbf{z}\|_2^2 + r\|\widehat{\mathbf{w}}\|_1 \leq \frac{1}{n_t}\|\boldsymbol{\Psi}^T\mathbf{w}^*-\mathbf{z}\|_2^2 + r\|\mathbf{w}^*\|_1.$$

Substituting $\mathbf{z} = \boldsymbol{\Psi}^T\mathbf{w}^* + \mathbf{e}$, one obtains

$$\frac{1}{n_t}\|\boldsymbol{\Psi}^T(\widehat{\mathbf{w}}-\mathbf{w}^*)-\mathbf{e}\|_2^2 \leq \frac{1}{n_t}\|\mathbf{e}\|_2^2 + r(\|\mathbf{w}^*\|_1-\|\widehat{\mathbf{w}}\|_1),$$

which, using the notations $\boldsymbol{\delta} = \widehat{\mathbf{w}} - \mathbf{w}^*$ and $r_e = \|\boldsymbol{\Psi}\mathbf{e}\|_\infty/n_t$, is expanded to give

$$\frac{1}{n_t}\|\boldsymbol{\Psi}^T\boldsymbol{\delta}\|_2^2 \leq \frac{2}{n_t}\boldsymbol{\delta}^T\boldsymbol{\Psi}\mathbf{e} + r(\|\mathbf{w}^*\|_1-\|\widehat{\mathbf{w}}\|_1),$$

$$\leq 2r_e\|\boldsymbol{\delta}\|_1 + r(\|\mathbf{w}^*\|_1 - \|\widehat{\mathbf{w}}\|_1),$$
$$= 2r_e(\|\boldsymbol{\delta}_J\|_1 + \|\widehat{\mathbf{w}}_{J^c}\|_1)$$
$$\quad + r(\|\mathbf{w}^*_J\|_1 - \|\widehat{\mathbf{w}}_J\|_1) - r\|\widehat{\mathbf{w}}_{J^c}\|_1,$$
$$\overset{(a)}{\leq} \|\boldsymbol{\delta}_J\|_1(2r_e+r) + \|\widehat{\mathbf{w}}_{J^c}\|_1(2r_e-r),$$
$$\leq \sqrt{s}\|\boldsymbol{\delta}_J\|_2(2r_e+r) + \|\widehat{\mathbf{w}}_{J^c}\|_1(2r_e-r), \quad (12)$$

where inequality (a) arises because $\|\mathbf{w}^*\|_1 - \|\widehat{\mathbf{w}}\|_1 \leq \|\mathbf{w}^* - \widehat{\mathbf{w}}\|_1 = \|\boldsymbol{\delta}_J\|_1$. Dividing both sides of (12) by $\|\boldsymbol{\Psi}^T\boldsymbol{\delta}\|_2$ gives

$$\frac{1}{n_t}\|\boldsymbol{\Psi}^T\boldsymbol{\delta}\|_2 \leq \frac{\sqrt{s}\|\boldsymbol{\delta}_J\|_2}{\|\boldsymbol{\Psi}^T\boldsymbol{\delta}\|_2}(2r_e+r) + \frac{\|\widehat{\mathbf{w}}_{J^c}\|_1}{\|\boldsymbol{\Psi}^T\boldsymbol{\delta}\|_2}(2r_e-r),$$

which is reduced to

$$\frac{1}{n_t}\|\boldsymbol{\Psi}^T\boldsymbol{\delta}\|_2 \leq 2r\sqrt{s}\frac{\|\boldsymbol{\delta}_J\|_2}{\|\boldsymbol{\Psi}^T\boldsymbol{\delta}\|_2}. \quad (13)$$

when $2r_e \leq r$. Clearly the inequality in (13) holds with probability no less than $P_e = p(2r_e \leq r)$. We will come back to find the expression of $P_e$; until then we assume $2r_e \leq r$ is true. We follow (Bickel et al., 2009; Lounici et al., 2009) to similarly define $\kappa_s = \min_{\boldsymbol{\delta}\neq\mathbf{0}} n_t^{-1/2}\|\boldsymbol{\delta}_J\|_2^{-1}\|\boldsymbol{\Psi}^T\boldsymbol{\delta}\|_2$, then

$$\|\boldsymbol{\delta}_J\|_2 \leq \kappa_s^{-1} n_t^{-1/2}\|\boldsymbol{\Psi}^T\boldsymbol{\delta}\|_2, \quad (14)$$

Substitution of (14) into (13) yields $\|\boldsymbol{\Psi}^T\boldsymbol{\delta}\|_2 \leq 2r\sqrt{n_t s}/\kappa_s$, which is substituted back into (14) to give

$$\|\boldsymbol{\delta}_J\|_2 \leq 2r\kappa_s^{-2}\sqrt{s}. \quad (15)$$

By the definition of $\kappa_s$, one has

$$n_t\kappa_s^2 = \min_{\boldsymbol{v}\neq\mathbf{0}}\frac{\|\boldsymbol{\Psi}^T\boldsymbol{v}\|_2^2}{\|\boldsymbol{v}_J\|_2^2} \geq \min_{\boldsymbol{v}\neq\mathbf{0}}\frac{\|\boldsymbol{\Psi}^T\boldsymbol{v}\|_2^2}{\|\boldsymbol{v}\|_2^2}.$$

Substituting (5), alongside (4), one gets

$$n_t\kappa_s^2 = \min_{\boldsymbol{v}\neq\mathbf{0}}\sum_{m=1}^M\frac{\|\mathbf{X}_m^T\mathbf{F}_m(\mathbf{F}_m^T\mathbf{F}_m)^{-1}\boldsymbol{v}\|_2^2}{\|\boldsymbol{v}\|_2^2},$$
$$\geq \sum_{m=1}^M \min_{\boldsymbol{v}\neq\mathbf{0}}\frac{\|\mathbf{X}_m^T\mathbf{F}_m(\mathbf{F}_m^T\mathbf{F}_m)^{-1}\boldsymbol{v}\|_2^2}{\|\boldsymbol{v}\|_2^2},$$
(Weyl's Inequality)

$$\begin{aligned}
&= \sum_{m=1}^M \min_{\boldsymbol{v}\neq\boldsymbol{0}} \frac{\|\mathbf{X}_m^T\mathbf{F}_m(\mathbf{F}_m^T\mathbf{F}_m)^{-1}\boldsymbol{v}\|_2^2}{\|\mathbf{F}_m(\mathbf{F}_m^T\mathbf{F}_m)^{-1}\boldsymbol{v}\|_2^2}\frac{\boldsymbol{v}^T(\mathbf{F}_m^T\mathbf{F}_m)^{-1}\boldsymbol{v}}{\boldsymbol{v}^T\boldsymbol{v}}, \\
&\geq \sum_{m=1}^M \min_{\boldsymbol{v}\neq\boldsymbol{0}} \frac{\|\mathbf{X}_m^T\mathbf{F}_m(\mathbf{F}_m^T\mathbf{F}_m)^{-1}\boldsymbol{v}\|_2^2}{\|\mathbf{F}_m(\mathbf{F}_m^T\mathbf{F}_m)^{-1}\boldsymbol{v}\|_2^2} \\
&\quad\times \min_{\boldsymbol{v}\neq\boldsymbol{0}}\frac{\boldsymbol{v}^T(\mathbf{F}_m^T\mathbf{F}_m)^{-1}\boldsymbol{v}}{\boldsymbol{v}^T\boldsymbol{v}}, \\
&\geq \sum_{m=1}^M \min_{\widetilde{\boldsymbol{v}}\neq\boldsymbol{0}}\frac{\|\mathbf{X}_m\widetilde{\boldsymbol{v}}\|_2^2}{\|\widetilde{\boldsymbol{v}}\|_2^2}\min_{\boldsymbol{v}\neq\boldsymbol{0}}\frac{\boldsymbol{v}^T(\mathbf{F}_m^T\mathbf{F}_m)^{-1}\boldsymbol{v}}{\boldsymbol{v}^T\boldsymbol{v}}, \\
&\geq \sum_{m=1}^M \frac{\omega_{\min}(\mathbf{X}_m^T\mathbf{X}_m)}{\omega_{\max}(\mathbf{F}_m^T\mathbf{F}_m)}, \quad (16)
\end{aligned}$$

where $\omega_{\min}(\cdot)$ and $\omega_{\max}(\cdot)$ respectively represents the maximum and minimum eigenvalue of a Hermitian matrix. Substitution of (16) into (15) gives

$$\|\boldsymbol{\delta}_J\|_2 \leq \frac{2r\sqrt{s}}{\sum_{m=1}^M \omega_{\min}(\mathbf{X}_m^T\mathbf{X}_m/n_t)\,\omega_{\max}^{-1}(\mathbf{F}_m^T\mathbf{F}_m)}. \quad (17)$$

By the result in (Byrne, 2009),

$$\omega_{max}(\mathbf{F}_m^T\mathbf{F}_m) \leq \max_i \left(\sum_{j=1}^{F_0} \|\mathbf{f}_{m,:,j}\|_0 |f_{ij}|^2\right),$$

$m = 1, 2, \cdots, M$, which is substituted into (17) to yield (8), using the auxiliary variable defined as $a = n_t r (\ln F_0)^{-1/2} \varepsilon_\psi^{-1}$ and $\|\boldsymbol{\delta}_{J^c}\|_2 \leq \|\boldsymbol{\delta}_{J^c}\|_1 \leq c_0\|\boldsymbol{\delta}_J\|_1 \leq c_0\sqrt{s}\|\boldsymbol{\delta}_J\|_2$, where $\|\boldsymbol{\delta}_{J^c}\|_1 \leq c_0\|\boldsymbol{\delta}_J\|_1$ by assumption.

Recall that (13) holds with probability no less than $P_e = p(2r_e \leq r)$. Since (8) is implied by (13), the probability for (8) being true is no less than $P_e$ also.

To evaluate $P_e$, we first plug $r_e = \|\boldsymbol{\Psi}\mathbf{e}\|_\infty/n_t$ into $P_e = p(2r_e \leq r)$ and expand the result, yielding

$$\begin{aligned}
P_e &= p(2\|\boldsymbol{\Psi}\mathbf{e}\|_\infty/n_t \leq r), \\
&= 1 - p(2\|\boldsymbol{\Psi}\mathbf{e}\|_\infty/n_t \geq r), \\
&\geq 1 - \sum_{j=1}^{F_0} p\left(2|\boldsymbol{\psi}_j^T\mathbf{e}|/n_t \geq r\right), \\
&= 1 - \sum_{j=1}^{F_0} p\left(|\boldsymbol{\psi}_j^T\mathbf{e}|\,\|\boldsymbol{\psi}_j\|_2^{-1} \geq n_t r 2^{-1}\|\boldsymbol{\psi}_j\|_2^{-1}\right), \\
&\geq 1 - \sum_{j=1}^{F_0} p\left(|\boldsymbol{\psi}_j^T\mathbf{e}|\,\|\boldsymbol{\psi}_j\|_2^{-1} \geq n_t r/(2\varepsilon_\psi)\right),
\end{aligned}$$

where the first inequality results from the union bound. Since the elements of $\mathbf{e}$ are i.i.d. from the standard normal distribution, so is $\boldsymbol{\psi}_j^T\mathbf{e}/\|\boldsymbol{\psi}_j\|_2$. Using the inequality $\mathbb{P}(|X| > x) \leq 2\exp\left(-\frac{x^2}{2}\right)/(x\sqrt{2\pi})$, $x > 0$, for any standard normal-distributed random number $X$, one obtains

$$\begin{aligned}
P_e &\geq 1 - \frac{4F_0 \exp\left(-\frac{n_t^2 r^2}{8\varepsilon_\psi^2}\right)}{\sqrt{2\pi}n_t r \varepsilon_\psi^{-1}} \\
&= 1 - \frac{4}{a\sqrt{2\pi \ln F_0}} F_0^{1-a^2/8}, \\
&\geq 1 - F_0^{1-a^2/8},
\end{aligned}$$

where the equation is due to $a = n_t r (\ln F_0)^{-1/2}\varepsilon_\psi^{-1}$, and the second inequality arises because $F_0 \geq 2$ and $a \geq \sqrt{8}$ by assumption, which ensure $\frac{4}{a\sqrt{2\pi \ln F_0}} \leq 1$. □